\documentclass[a4paper,fleqn,colorlinks,breaklinks=true,pdftex]{cas-sc}
\usepackage[authoryear,longnamesfirst]{natbib}

\usepackage{lineno}
\usepackage{graphicx}	
\usepackage{wrapfig}
\usepackage{multirow}
\usepackage{amsmath}
\usepackage{color}
\usepackage{graphics} 
\usepackage{overpic}
\usepackage{epsfig} 
\usepackage{amsmath} 
\usepackage{amssymb} 

\usepackage{hyperref}

\graphicspath{{figures/images/}}

\def\tsc#1{\csdef{#1}{\textsc{\lowercase{#1}}\xspace}}
\tsc{WGM}
\tsc{QE}
\tsc{EP}
\tsc{PMS}
\tsc{BEC}
\tsc{DE}

\begin{document}
\let\WriteBookmarks\relax
\def\floatpagepagefraction{1}
\def\textpagefraction{.001}
\shorttitle{Physics-Aware Iterative Learning and Prediction of Saliency Map for Bimanual Grasp Planning}
\shortauthors{SY Wang et~al.}

\title [mode = title]{Physics-Aware Iterative Learning and Prediction of Saliency Map for Bimanual Grasp Planning}
\author[1]{Shiyao Wang}[style=english]
\ead{shiyaowang@mail.dlut.edu.cn}
\author[1]{Xiuping Liu}[style=english]
\ead{xpliu@dlut.edu.cn}
\author[2]{Charlie C.L. Wang}[style=english]
\ead{changling.wang@manchester.ac.uk}
\author[3]{Jian Liu}[style=english]
\ead{jianliu2006@gmail.com}
\cormark[1]

\address[1]{School of Mathematics Sciences, Dalian University of Technology, Dalian 116024}
\address[2]{School of Engineering, The University of Manchester, United Kingdom}
\address[3]{School of Software, Tsinghua University, Beijing 100084, P. R. China}

\cormark[1]

\cortext[cor1]{Corresponding author}

\begin{abstract}
Learning the skill of human bimanual grasping can extend the capabilities of robotic systems when grasping large or heavy objects. However, it requires a much larger search space for grasp points than single-hand grasping and numerous bimanual grasping annotations for network learning, making both data-driven or analytical grasping methods inefficient and insufficient. We propose a framework for bimanual grasp saliency learning that aims to predict the contact points for bimanual grasping based on existing human single-handed grasping data. We learn saliency corresponding vectors through minimal bimanual contact annotations that establishes correspondences between grasp positions of both hands, capable of eliminating the need for training a large-scale bimanual grasp dataset. The existing single-handed grasp saliency value serves as the initial value for bimanual grasp saliency, and we learn a \emph{saliency adjusted score} that adds the initial value to obtain the final bimanual grasp saliency value, capable of predicting preferred bimanual grasp positions from single-handed grasp saliency. We also introduce a physics-balance loss function and a physics-aware refinement module that enables physical grasp balance, capable of enhancing the generalization of unknown objects. Comprehensive experiments in simulation and comparisons on dexterous grippers have demonstrated that our method can achieve balanced bimanual grasping effectively.
\end{abstract}

\maketitle

\section{Introduction}\label{sec:intro}
In manual operation environments, bimanual interactions with household objects are required for many daily tasks. In comparison to the single-handed grasping, grasping with both hands has obvious advantages in terms of stability and balance for big or heavy objects. For example, when grasping an empty kettle, people prefer to grasp the handle with a single hand. However, when the kettle is full of water, people typically choose to grasp it with both hands, which is more stable than grasping it with only one hand. Learning from this bimanual grasping experience can significantly extend the capability of bimanual robotic system for completing the tasks of bimanual grasping in similar scenarios.

Although there are many works on robotic grasping, the problem of bimanual grasp planning is still a challenge. Planning a feasible bimanual grasp needs to search contact points of both hands on the object surface using physics-based analyses such as force-closure~\cite{Dai-2015SynthesisAO}, which results in a search space much larger than that for contact points of single hand. To solve this problem, previous works conduct the bimanual grasping either by regarding two hands of the robot as two fingers connected to a wrist and then solve it with single-grasp planners~\cite{Vahrenkamp-2011BimanualGP,Berenson-2008GraspSI,Caraza-1991TwohandedGW}, or separately computing potential contact points of each hand for grasp candidate selection~\cite{Seo-2012SpatialBW,Saut-2010PlanningPT,Przybylski-2011PlanningGF,Hayakawa-9590084}. Furthermore, to compute a grasp using both hands simultaneously,~\cite{RojasdeSilva-2016GraspingBO,RojasdeSilva-2017ContactFC} approaches split the object surface into two slices from which the contact points of each hand are selected. While these methods strive to narrow down the search space, they overly reduce the feasibility of choosing more stable bimanual contact points. 

On the other aspect, human demonstrations have been exploited to improve robotic grasping and 3D grasp synthesis in data-driven methods~\cite{Amor-2012GeneralizationOH,Yu-9180002,Zhu-2021TowardHG,Brahmbhatt-2020ContactPoseAD,Song-7051290}. However, it is challenging to collect a large number of human grasp trials that requires a significant cost. Some recent works replace human grasp demonstration with semantic hand-object contact representation such as contact map~\cite{Brahmbhatt-2019ContactDBAA}, saliency map~\cite{Lau-2016TactileMS,Zhang-20203DGS} and affordence map~\cite{Mandikal-2021LearningDG,Deng-20213DAA,Xu-2022PartAffordPA,Song-7051290} to synthesize the grasps. ContactGrasp~\cite{Brahmbhatt-2019ContactGraspFM} involved robotic grasp synthesis for dexterous hands from the contact maps recorded by ContactDB~\cite{Brahmbhatt-2019ContactDBAA}. \cite{Lau-2016TactileMS} collected the dataset of human grasping preference through crow-sourcing and utilized a deep ranking framework to predict grasp saliency map. However, these methods can't directly address the challenging problem of bimanual grasping since they mainly focus on the single-handed grasping task and no existing large-scale dataset of human annotations of bimanual contact points. These defects raise a challenge for the network to predict specific contact areas that humans prefer to grasp with both hands in the absence of adequate training data.

\begin{figure}[t!]
  \centering
  \begin{overpic}[width=1\columnwidth,tics=10]{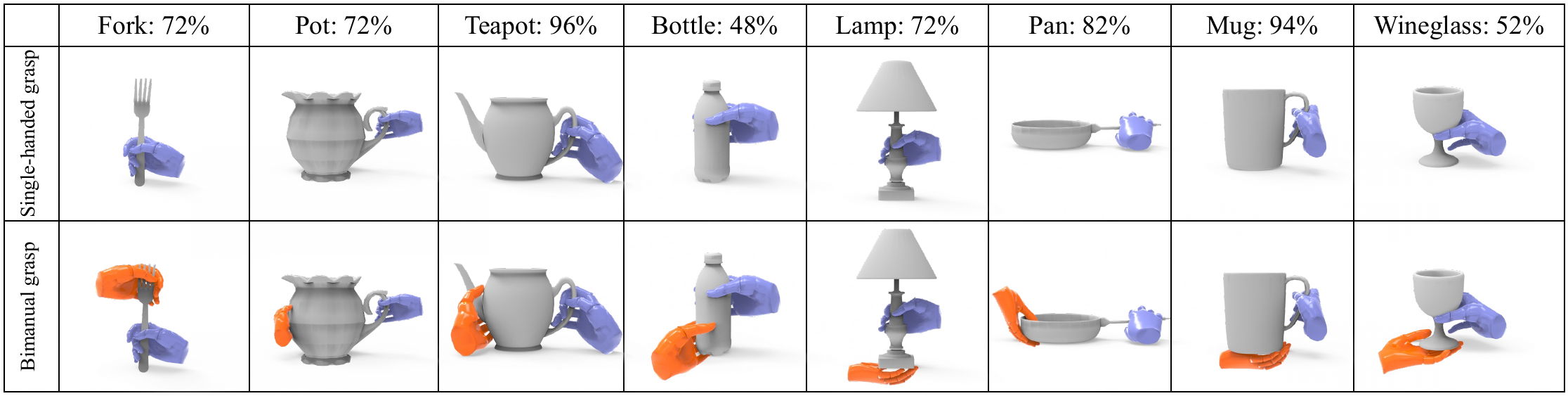}
  \end{overpic}

  \caption{The figure shows the average grasp coverage of 8 categories of objects and examples of single-handed grasp poses and bimanual grasp poses. The grasp coverage is defined as the percentage of users who labeled single-handed grasping positions are covered by bimanual grasping positions.}
   \label{fig:user_study}
   \vspace{-7pt}
\end{figure}

Based on the observation of human grasping behavior, two intuitive correlations between single-handed grasping and bimanual grasping we founded may help to tackle the challenge. To illustrate the correlation, we invited 30 participants and asked them to annotate grasps for 8 categories of household objects with only one and both hands, respectively. As shown in figure~\ref{fig:user_study}, humans usually choose bimanual contact points based on object’s geometry and affordence. It can be observed that the contact position of right hand is generally located in the functional area of object, while the corresponding contact position of left hand would be selected to assist in grasping balance or support. We further compute the grasp coverage for each object, which is the degree of overlap between bimanual contact areas and single-handed contact areas. It shows that the grasp coverage is above 70\% for most categories of objects, indicating that humans usually still choose the specific grasping area similar to single-hand grasping when conducting bimanual grasping. 

Inspired by the aforementioned intuitive correspondence and similarity between single-handed grasping and bimanual grasping, we advocate to predict bimanual contact points relying on human single-handed grasping preference, rather than relying on large-scale data with bimanual contact annotations. To this end, we propose a bimanual grasp saliency learning framework for bimanual contact point detection. The existing human single-handed grasp saliency map is served as initiation of bimanual saliency map, and we build a small number of manual annotations of bimanual contact points to learn correspondence between grasp positions of both hands. Therefore, our method learns saliency adjusted score that adjusts the initial value to bimanual saliency value. In addition, to make the initial value of bimanual saliency be fine-tuned, we also design an iterative learning strategy so that the initial value can be updated. In order to further achieve the generalization when human single-handed grasping experience fail in bimanual grasping for unseen objects that are not included in the training dataset since the lack of large-scale dataset, our method takes the physical factors of grasping into account for both training and inference stages. 
Finally, we train a segmentation network to predict bimanual contact points from the bimanual grasp saliency map and generate bimanual grasp poses relying on the existing 3D grasp synthesis approach~\cite{Brahmbhatt-2019ContactGraspFM}.

Given an input object point cloud, we propose a physics-aware bimanual saliency iterative learning framework for bimanual grasp planning. 
Our method does not rely on a large number of bimanual contact annotation, but only adjusts the single-handed saliency map to the bimanual saliency map through minimal human annotations and physical criterion, so as to obtain the contact points of both hands. 

We conducted comprehensive experiments on a variety of household objects and evaluated our approach in simulation. Results have demonstrated that our method can generate stable grasps with predicted bimanual contact points effectively. In summary, our work makes the following contributions:
\begin{itemize}
\item We exploit the correlations between single-handed grasping and bimanual grasping and propose a neural network-based method to predict the bimanual grasp saliency map only relying on existing human single-handed grasping preference saliency data and a small dataset of manual annotations.
\item We propose a bimanual saliency iterative learning strategy so that the initial value of bimanual saliency can be efficiently updated during both training and inference stages.
\item We take the constraint of physical balance in grasping into account and design a physical balance based loss function and a physics-aware refinement to enhance the generalization for bimanual grasping of objects that have not been `seen' in the training dataset.

\end{itemize}
\section{Related work} \label{sec:relatedwork}
\subsection{Bimanual grasp Planners.} Planning bimanual grasp is an important field of robotic grasping in which several studies and practical applications have been found~\cite{Smith-2012DualAM}. Many relevant works focus on the force-closure methods to search contact points on the object surface and synthesize feasible grasps of both hands. To plan the bimanual grasping, \cite{Berenson-2008GraspSI} formalized both hands of the robot as an abstraction of two fingers connected to a wrist and converted the bimanual grasping problem into a single-handed grasping problem. \cite{Seo-2012SpatialBW} used the whole body to grasp polyhedral objects where both arms are represented by chain grippers and the feasible contact configurations are selected. \cite{Vahrenkamp-2009HumanoidMP} proposed to address the bimanual grasping problem by choosing grasp candidates for each hand separately. To avoid suffering from manipulability of bimanual grasp, the approach~\cite{Vahrenkamp-2011BimanualGP} computed the reachable grasp candidates of each end-effector by inverse kinematics and then found force-closure grasp configurations performed by the grasp wrench space analysis. Unlike the aforementioned discussed works, \cite{RojasdeSilva-2016GraspingBO} computed a kinematically reachable grasp of two dexterous hands simultaneously. In this process, the object observed as the point cloud is divided into two parts where the contact points of both hands are selected, and then the combinations of the two-handed contact points are evaluated to satisfy the force closure condition. After potential contact points are determined, the suitable grasp configuration is performed by computing inverse kinematics and checking collision. Although these grasp planners can be used to compute bimanual grasps by simplifying the search space of bimanual contacts, they overly reduce the feasibility of choosing more stable bimanual contact points. In contrast to these previous works, we begin to shift focus towards using human grasp preference to plan contact points for bimanual grasp, which helps in efficiently finding feasible bimanual grasps.

\subsection{Grasp synthesis from hand-object contact.} 
The research of grasp synthesis recently focus on the semantic hand-object representation (contact map, grasp saliency, grasp affordance, etc.) to synthesize task-oriented grasp poses of robot grippers. \cite{Brahmbhatt-2019ContactDBAA} first annotated contact maps on the object surfaces that represent human-like functional grasp regions of the human hand. Based on that, ContactGrasp~\cite{Brahmbhatt-2019ContactGraspFM} synthesized functional robotic grasps by optimizing stable grasp configurations computed by a state-of-art grasp planner to align with the contact maps. ContactPose~\cite{Brahmbhatt-2020ContactPoseAD} further recorded both human hand poses and contact points to learn functional grasps of the dexterous hands. Some relevant works also used contact map as an intermediate variable to constrain the grasp generation~\cite{li2023contact2grasp} or to refine generated grasps during inference~\cite{Jiang2021HandObjectCC}. However, different grasps would sometimes result in the same contact map because of each part of the hand does not correspond to the specific part of the contact map. To overcome this issue caused by the contact map, \cite{Zhu-2021TowardHG} present a high-level semantic touch code that described which part of the hand contact the object that are manually pre-segmented into the functional parts for grasping and train a deep architecture to generate human-like grasps~\cite{Wu2023FunctionalGT,Zhang2023FunctionalGraspLF}. In contrast to the contact map represented from human demonstration, the semantic hand-object representation based on human visual perception is also widely used for grasp detection. \cite{Chen2022Learning6T} present an affordance-based network to predict grasp affordance map given an input point cloud and select candidate grasps based on it. In addition, \cite{Lau-2016TactileMS} first proposed to predict grasp saliency, which is an important tool to explore human grasp skills. This approach first defined the grasp salient points on the mesh model that human would like to grasp and then instructed online user annotators to crowd-source such saliency data. After that, a deep ranking technique is developed to measure grasp saliency. However, the grasp saliency prediction heavily relies on sufficient grasp annotations, which requires an expensive acquisition cost. To overcome this challenge, \cite{Zhang-20203DGS} developed a transfer learning methodology to augment saliency data via deep shape correspondence. While these approaches contribute to robotic analyses of task-oriented human-like grasping, they cannot directly specify bimanual grasping, only relying on  human grasp data of the single hand. Different from these methods, our solution is able to predict the bimanual grasp saliency map for bimanual grasp planning relying on human single-handed grasping preference without depending on large-scale data with bimanual contact annotations.
\begin{figure}[ht]
  \centering
  \begin{overpic}[width=1.0\columnwidth,tics=10]{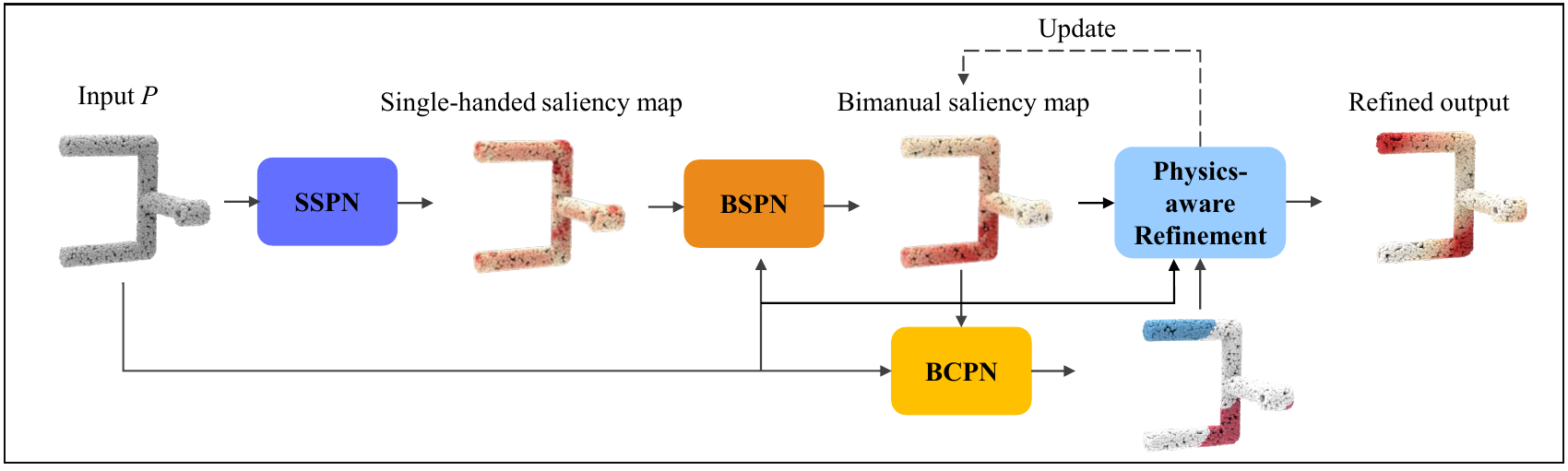}
  \end{overpic}
  \caption{
The overview of our physics-aware iterative learning and prediction of bimanual saliency map for novel objects. We first use a pre-trained SSPN model to compute the single-handed saliency map given an input point cloud. Next, the proposed BSPN network is leveraged to predict bimanual saliency map, which takes both point cloud and the single-handed saliency map, and the bimanual contact points are computed by applying BCPN network given both point cloud and the result of BSPN. After that, we refine bimanual saliency map to conform the physical stability using physics-aware refinement module.
}
   \label{fig:pipeline}
   \vspace{-7pt}
\end{figure} 
\section{Method}\label{sec:method}
This work concentrates on predicting a physics-aware saliency map of object for bimanual grasp planning, which implies predicting bimanual grasp contact points that are physically plausible for grasping balance and amenable to human grasping preferences. In order to predict bimanual saliency map in absence of training data, we propose a novel framework that exploits physics-aware bimanual saliency map from the single-handed saliency map with minimal bimanual grasp contact annotations, as shown in Figure~\ref{fig:pipeline}. It generates maps for bimanual saliency grasp areas by the bimanual saliency map prediction network (Sec.~\ref{sec:bspn}), named as BSPN, computes the contact points of both hands from bimanual saliency map by the bimanual contact point prediction network (Sec.~\ref{sec:contactPointPrediction}), named as BCPN, refines the predicted result of bimanual saliency map by the physics-aware refinement (Sec.~\ref{sec:refinement}), and synthesizes grasp configurations from computed contact points by ContactGrasp~\cite{Brahmbhatt-2019ContactGraspFM}, a grasp optimizer that computes grasp poses (Sec.~\ref{sec:Application}).

\subsection{Problem Statement}
Given an input object point cloud $P=\{x_{i}\}^{N}_{i=1}$, we define its bimanual saliency map $B=\{b_i\}^{N}_{i=1}$ and single-handed saliency map $S=\{s_i\}^{N}_{i=1}$, where $b_i \in [0,1]$ and $s_i \in [0,1]$ denote the bimanual saliency value and single-handed saliency value of the point $x_i$ respectively. The closer the saliency value of point $x_i$ is to 1, the more likely it is to be grasped. Moreover, to annotate bimanual grasp contact points, we also define $L=\{l_{i}\}^{N}_{i=1}$ as a set of bimanual-contact labels, where $l_i$ is formula as follow:
\begin{equation*}
l_{i} =\begin{cases}
1 & if\ x_{i} \ \in \ right-hand, \\
2 & elif\ x_{i} \ \in \ left-hand, \\
0 & otherwise.
\end{cases}
\end{equation*}
$x_{i} \in \ right-hand$ denotes that $x_i$ is annotated as the grasp contact point of right hand and $x_{i} \in \ left-hand$ denotes that $x_i$ is annotated as the grasp contact point of left hand. More details about building the bimanual-contact label set is described in Sec.~\ref{sec:Implementation}. Notice that we only use a small number of bimanual annotations in our approach, as the extensive annotation is time-consuming.

Based on the observation of human grasping behavior with both hands, we discover that two intuitive correlations between single-handed grasping and bimanual grasping. Firstly, the contact position of the right hand is generally located in the functional area of the object, while the corresponding contact position of the left hand would be selected to assist in grasping balance and support. Secondly, the grasp contacts chosen by one hand are covered by the grasp contacts of both hands. These correlations give us an important insight about predicting bimanual saliency map by taking the single-handed saliency map as initialization. Motivated by this, we compute the bimanual saliency map $B$ using single-handed saliency map $S$ and the spatial relationship between the grasping positions of both hands. We define the relationships as correspondences between two kinds of grasping areas: the correspondences between the points labeled \textit{left-hand} and the points labeled \textit{right-hand}. Specifically, for a pair of points $x_{i}$ and $x_{j}$ that are annotated as \textit{left-hand} and \textit{right-hand} respectively, we define their saliency corresponding vector $v_{ij} \subset V$ as a displacement from $x_{i}$ to $x_{j}$. Thus, the corresponding relationship could be presented as:
\begin{equation}
  x_{j}=x_{i}+v_{ij}.
\end{equation}
Our basic idea is, for a pair of points ($x_{i}$,$x_{j})$ that are annotated as \textit{left-hand} and \textit{right-hand} respectively, their bimanual saliency values $b_{i}$ and $b_{j}$ should be as similar as possible. Therefore, we aim to find a mapping function $F_{\theta}:(S|P) \rightarrowtail \Delta S$ to predict a saliency adjusted score $\Delta S \in R^{N \times 1}$ which adjusts single-handed saliency map to bimanual saliency map based on the saliency corresponding vector. Thus, the bimanual saliency map can be formulated as $B =S +\Delta S$. In particular, learning the adjusted score should not only make the bimanual saliency map amenable to the human grasping perception, but also satisfy the physical balance and stability for grasping. 

Although we can obtain the bimanual saliency map by adjusting the single-handed saliency map, it may not adjust the saliency value of both hands simultaneously if we only rely on a fixed single-handed saliency map, which is due to the bimanual saliency value of the points labeled \textit{right-hand} are unchanged during the process. Hence, we develop an iterative learning strategy to update the single-handed saliency map.

In order to further ensure the generalization for novel objects when human single-handed grasping experience fail in bimanual grasping, we also take the grasping balance into account for both training and inference stages. On the one hand, we add the physics-balance loss function to bimanual saliency map learning. On the other hand, we design a physics-aware refinement module to refine the predicted result of the bimanual saliency map so that it conforms to physical stability.
\begin{figure}[t!]
  \centering
  \begin{overpic}[width=1\columnwidth,tics=10]{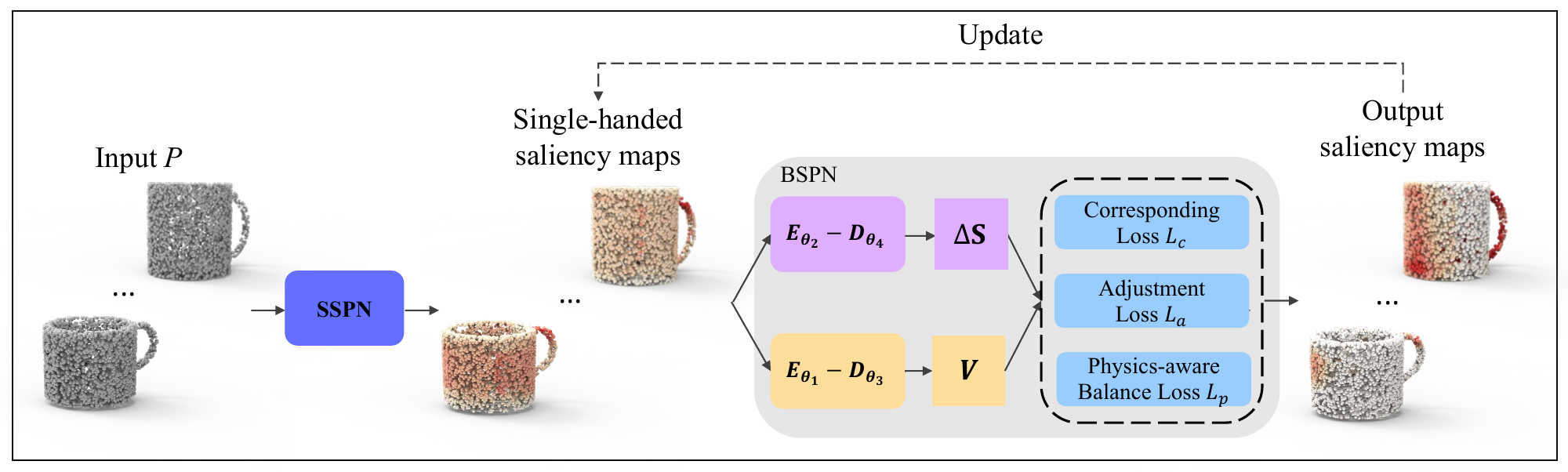}
  \end{overpic}
  \caption{The network architecture of  BSPN. The encoder is consist of 3-layer MLP. And the decoder is consist of 4-layer MLP, which takes the feature concatenated by 64-dimensional and 1024-dimensional features as input. Finally, the decoder outputs $m-$dimensional prediction. For the saliency corresponding vector $V$ and the saliency adjusted score $\Delta S$, $m$ is $3$ and $1$ respectively. The total training loss function consists of three items: Correspondence Loss $L_c$, Adjustment Loss $L_a$ and Physics-aware Balance Loss $L_p$.}
   \label{fig:BSPN_subnetwork}
   \vspace{-7pt}
\end{figure}
\subsection{Bimanual Saliency Map Prediction Network}
\label{sec:bspn}
Figure~\ref{fig:BSPN_subnetwork} demonstrates the architecture of BSPN, which takes both an object point cloud $P$ and the single-handed saliency map $S$ as inputs and outputs the bimanual saliency map $B$. We first generate the single-handed saliency map as the initialization of the bimanual saliency map. Then, we adjust the single-handed saliency map based on the bimanual grasp contact annotations and the physical criterion to generate a bimanual saliency map. Finally, we also propose an iterative learning strategy that updates the single-handed saliency map both in the training and inference stages. Next, we introduce the single-handed saliency map generation, the network structure and the iterative learning strategy of BSPN.

\subsubsection{Single-handed saliency map}
The single-handed saliency map is generated based on SSPN~\cite{Zhang-20203DGS} which takes a local patch centered on a point $x_i$ as input to predict the single-handed saliency value $s_i^o$ of $x_i$. We take the patch for each point as input to SSPN, then traverse all points of the point cloud and finally output the single-handed saliency map $S^{o}$. 

It is worth notice that since the SSPN is pre-trained on the existing training data of single-handed grasp saliency, the single-handed saliency areas predicted by SSPN may sometimes incompletely coincide with the position of points labeled \textit{right-hand} in our bimanual grasp contact annotation. Therefore, towards addressing this issue, we design and train a correction module (CM) for correcting the single-handed saliency map generated by SSPN. Specifically, it takes both an object point cloud and the single-handed saliency map $S^{o}$ as inputs and outputs the corrected single-handed saliency map $S^{c}$. Thus, the single-handed saliency map after correction can be written as $S=S^{o}+S^{c}$. The CM network also follows encoder-decoder architecture that is built on PointNet. In the training stage, we force the single-handed saliency value of points labeled \textit{right-hand} to be more saliency by making $L_{correct}$ close to 0. The $L_{correct}$ is formulated as follow:
\begin{equation}
L_{correct}=\frac{1}{N}(\sum_{i=1}^{N} E_{g}(x_{i})\cdot (s_{i}-1)^{2}+\sum_{i=1}^{N}E_{n}(x_{i})\cdot s_{i}^{2}),
\end{equation}
where $\displaystyle E_{g}(x_{i}) =\begin{cases}
1 & if\ l_{i}\ = 1,\\
0 & otherwise.
\end{cases}$ and $\displaystyle E_{n}(x_{i}) =\begin{cases}
1 & if\ l_{i}\ \in \{0,2\}\\
0 & otherwise.
\end{cases}$.
Hence, for points labeled \textit{right-hand}, we encourage their single-handed saliency values to be close to 1, and for other points, we encourage their single-handed saliency values to be close to 0. After obtaining the single-handed saliency map, we take it as the initial value of the bimanual saliency map. 

\subsubsection{Network structure}
As shown in Figure~\ref{fig:BSPN_subnetwork}, the BSPN is a supervised network, which is composed of two encoder-decoder architectures. Both two encoders $E_{\theta_1}:(P) \rightarrowtail f_t$ and $E_{\theta_2}:(P) \rightarrowtail f_s$ are built on PointNet~\cite{qi2016pointnet} architecture. The encoder $E_{\theta_1}$ takes a point cloud as input to extract the spatial location feature $f_s \in \mathbb{R}^{N \times 1024}$ associated with the saliency corresponding vectors. The encoder $E_{\theta_2}$ takes a point cloud as input to extract the saliency transformation feature $f_t \in \mathbb{R}^{N \times 1024}$ associated with the saliency adjusted score. For each encoder, according to the PointNet architecture, the feature representation for each point is computed by 5-layer MLP architecture. The features of each point are then aggregated into the global feature representation of the input point cloud by max-pooling. After that, we replicate the global feature $N$ times as a global feature representation for each point. Considering adding local geometric information, we need to concatenate the features $f_s^1$ and $f_t^1$ of the first layer of MLP architecture with the features output from the last layer, respectively.

Both decoders $D_{\theta_3}:(f_t,f_t^1) \rightarrowtail V$ and $D_{\theta_4}:(f_s,f_s^1) \rightarrowtail \Delta S$ are regressors to predict the saliency corresponding vectors and the saliency adjusted score, respectively. They use 4-layer MLP architecture. The decoder $D_{\theta_3}$ takes both the saliency transformation features $f_t$ and $f_t^1$ as inputs and outputs the saliency corresponding vectors $V$. The decoder $D_{\theta_4}$ takes both the spatial location features $f_s$ and $f_s^1$ as inputs and outputs the saliency adjusted score $\Delta S$. In the inference stage, only $E_{\theta_2}$ and $D_{\theta_4}$ are used. The encoder $E_{\theta_2}$ takes point cloud $P$ to output the spatial location features. With these features, the decoder $D_{\theta_4}$ outputs the saliency adjusted score $\Delta S$, and then we generate the bimanual saliency map by adding the saliency adjusted score to the single-handed saliency map.

\subsubsection{Loss function}
To enable our BSPN network to generate reasonable bimanual saliency map by adjusting the single-handed saliency, we design the loss functions based on the single-handed saliency map, bimanual grasp contact annotation and physical criterion. First, we use two loss functions to constrain the saliency corresponding vectors $V$ and saliency adjusted score $\Delta S$. The first is the correspondence loss $L_c$, which is used to constrain the spatial correspondences between the points labeled \textit{left-hand} and the points labeled \textit{right-hand}. The second is the adjustment loss $L_{a}$, which makes the adjustment of the single-handed saliency map well. Besides that, to ensure that the generated bimanual saliency map is physically plausible for grasp balance, we also propose a physics-balance loss $L_{p}$. Overall, the total loss function consists of three items: correspondence loss, adjustment loss and physics-balance loss.

\emph{\textbf{Correspondence Loss}} The correspondence loss makes the errors between the predicted corresponding vectors and ground truth corresponding vectors as small as possible. For a given predicted saliency corresponding vector $v_i$ and the ground truth $\hat{v_{i}}$, the loss can be formulated as the mean squared error between $v_i$ and $\hat{v_{i}}$:
\begin{equation}
L_c=\frac{1}{N}(\sum_{i=1}^{N} \phi(x_{i})\cdot(v_{i} -\hat{v_{i}})^{2}),
\end{equation}
where $\phi(x_{i}) =\begin{cases}
1 & if\ l_{i}\ \in \ \{1,2\},\\
0 & otherwise.
\end{cases}$
$\phi(x_i)$ is an indicative function. If $x_i$ is annotated as either \textit{left-hand} or \textit{right-hand}, $\phi(x_i)$ is equal to $1$ otherwise $\phi(x_i)$ is equal to $0$. This means that we only need to focus on the accuracy of the saliency corresponding vector prediction of points in annotation.

\emph{\textbf{Adjustment Loss}} The points labeled \textit{left-hand} and the points labeled \textit{right-hand} are the potential contact points of bimanual grasping, which means that their bimanual saliency values seem to be almost identical. Therefore, we use corresponding vector to constrain bimaunal saliency values of a pair of points labeled \textit{left-hand} and labeled \textit{right-hand} to be as similar as possible. Given $b_{i}$ and $b_{j}$, the goal is achieved by optimizing the mean squared error:
\begin{equation}
L_{a_1}=\frac{1}{N}(\sum_{i=1}^{N} \phi(x_{i})\cdot( b_{i} -b_{j})^{2}).
\end{equation}
Furthermore, for the points labeled \textit{right-hand} and points labeled neither \textit{left-hand} nor \textit{right-hand}, that is, the points with $l_i \in \{0,1\}$, we believe that their bimanual saliency values do not need to change in the adjustment process. Their bimaunal saliency values should be consistent with the single-handed saliency value. Given the bimanual saliency value $b_i$ and the single-handed saliency value $s_i$, the loss formula is as follow:
\begin{equation}
L_{a_2}=\frac{1}{N}(\sum_{i=1}^{N} \psi(x_{i})\cdot(b_{i}-s_{i})^{2}),
\end{equation}
where $\displaystyle \psi(x_{i}) =\begin{cases}
1 & if\ l_{i}\ \in \{0,1\},\\
0 & otherwise.
\end{cases}$. $\psi(x)$ is also an indicator. 
Therefore, the adjustment loss function $L_a$ can be written as:
\begin{equation}
L_{a}=\lambda_{1} \cdot L_{a_1}+\lambda_{2} \cdot L_{a_2},
\end{equation}
where $\lambda_{1}=1$ and $\lambda_{2}=1.5$ is the corresponding weights.

\emph{\textbf{Physics-balance Loss}} We force the midpoint between the left-hand contact point and the right-hand contact point to approach the direction of gravity, which ensures the physical balance for grasping guided by the bimanual saliency map. Specifically, we select a pair of contact points $x_l$ and $x_r$ with the highest bimanual saliency value among the points labeled \textit{left-hand} and the points labeled \textit{right-hand}. Assuming the mass of the object is uniform, we set the object's geometric center $p_{center}$ as the center of gravity and select the vertical downward direction as the direction of gravity. $x_c$ is the midpoint between $x_l$ and $x_r$, and we calculate the Euclidean distance from the relative coordinates of midpoint $x_c-p_{center}$ to the straight line $G$ where the gravity direction is located, and set it as the physical balance loss, which could be presented as:
\begin{equation}\label{eq:physics-balance loss}
L_{p}=D(x_c-p_{center},G),
\end{equation}
where $D(\cdot,\cdot)$ is the Euclidean distance from $x_c-p_{center}$ to the straight line $G$. 

The final loss function is the linearly combination of the correspondence item, the adjustment item and the physics-balance loss item, which could be presented as:
\begin{equation}
L_{total}=w_1 \cdot L_c+w_2 \cdot L_{a}+w_3 \cdot L_p.
\end{equation}
where $w_1=1$, $w_2=1$ and $w_3=2$ denote the corresponding weights.

\subsubsection{Iterative Learning Strategy}
\label{sec:Iterative Training Strategy}
We also need to update the single-handed grasp saliency map in the adjustment process. Particularly, for the points annotated as \textit{right-hand}, their single-handed saliency values need to change since we use a pre-generated single-handed saliency map as initialization. To this end, we employ an iterative learning strategy in BSPN. Specifically, we update the single-handed saliency map $S$ when the training epoch $>=K$ and every $M$ epochs. It updates $m$ times in total, which can be formula as $S^{t}=B^{t-1}, t=1,2,\cdots,m$, where $B^{0}$ is equal to the initial single-handed saliency map. Meanwhile, we also need to re-formulate the adjustment loss $L_{a_2}$ since the original item but only constrain the points labeled \textit{right-hand} and but also constrain the points labeled neither \textit{left-hand} nor \textit{right-hand}. In the iteration phase, we only need to force $L_{a_2}$ to constrain the points labeled neither \textit{left-hand} nor \textit{right-hand} by representing the indicator $\psi(x)$ as:
$\displaystyle \psi(x_{i}) =\begin{cases}
1 & if\ l_{i}\ = 0,\\
0 & otherwise.
\end{cases}$.

Besides that, we also need to give a stop condition for the iteration. Instead of terminating the iteration by manually setting the iteration number, which is hard to control the reasonable of results, we design an automatic mechanism for iteration termination based on the generated bimanual saliency map and the physics-balance loss function. Specifically, we define two stop conditions. One condition is the average bimanual saliency value of all the points labeled \textit{left-hand} and the points labeled \textit{right-hand}. The other condition is the physics-balance distance that is used in Eq~\ref{eq:physics-balance loss}. Therefore, the iteration is automatically terminated when the average bimanual saliency value reaches to a threshold $\sigma_s$ and the physics-balance distance is less than a threshold $\sigma_p$ during iteration. In practice, $\sigma_s$ and $\sigma_p$ are set as $0.8$ and $0.12$, respectively.

\subsection{Bimanual Contact Point Prediction Network}
\label{sec:contactPointPrediction}
We compute bimanual contact points based on the object point cloud, predicted bimanual saliency map and the bimanual contact label. We develop a neural network, named bimanual contact point prediction network (BCPN), to compute the bimanual contact points.

The BCPN is built on the PointNet segmentation network which takes the concatenation of the point cloud and predicted bimanual saliency map as input and outputs the bimanual contact labels. We used it as a classifier $C_{\theta}:(B,P) \rightarrowtail L$ for every point $x_i \in P$ to classify whether it labels \textit{right-hand}, \textit{left-hand} or not. A small number of bimanual-contact labels we captured is represented as the ground truth for training. To classify each point, we adopt the cross-entropy loss function. We train the BCPN with the BSPN architecture together, which is due to the prediction of the saliency corresponding vector in BSPN and the classification of contact points both require the geometric information of the bimanual grasping position. Therefore, we an additional BSPN branch to assist the training of BCPN. The final loss function is defined as:
\begin{equation}
L_{classify}=w_4 \cdot\frac{1}{N}(-\sum_{i=1}^{N} \hat{l_i}log(l_i))+L_{total},
\end{equation}
where $l_i$ is the predicted bimanual contact label, $\hat{l_i}$ is the ground truth of the bimanual contact label and $w_4=1.5$ is constant for balancing the loss terms. Note that the BSPN is only employed in training and discarded in inference.

After predicting the bimanual contact labels, we compute bimanual contact points by combing with the bimanual saliency map. Specifically, we threshold bimanual saliency map at $\tau$ and define a bimanual saliency mask: $\displaystyle m_i =\begin{cases}
1 & if\ b_{i} >= \tau,\\
0 & otherwise. \end{cases}$. Therefore, we can use $l_i=l_i \cdot m_i$ to classify whether $x_i$ is a contact point or not. 

\subsection{Physics-aware Refinement}
\label{sec:refinement}
Although the BSPN network works well for the objects of the same category as the training data, it is not generalized enough for novel objects since the small amount of training data. Hence, we propose a physics-aware refinement module to further ensure generalization ability across objects. 

The physics-aware refinement aims to refine the predicted bimanual saliency map to conform to the physical stability. Specifically, the refinement module takes an object point cloud, bimanual saliency map $B$ predicted by BSPN and bimanual contact labels predicted by BCPN as input and utilizes the MLP architecture to compute the refined adjustment $R$ of per point’s bimanual saliency value. The refined bimanual saliency map is formula as: $B_r=B+R$. We hope that a pair of the grasp contact points, having the highest refined bimanual saliency value, of which one is labeled~\textit{right-hand} and the other is not labeled~\textit{right-hand}, conforms to the physical stability. Therefore, the distance between their midpoint and the straight line where the gravity direction is located should be as small as possible. The objective function for the physics-aware refinement is similar with the Physics-balance loss function Eq~\ref{eq:physics-balance loss}. We iteratively refine the bimanual saliency map until the objective function is smaller than the threshold $w_r$, where $w_r=0.12$.

To compute the final bimanual contact points, we concatenate the point cloud and the refined bimanual saliency map as input and apply K-means clustering~\cite{krishna1999genetic} based on the concatenation. We set $K=3$ and the points are divided into three categories: left-hand contact point, right-hand contact point and others. In this way, we can obtain the contact points of both hands, which could be used to plan the bimanual grasping.

\subsection{Application: bimanual grasp synthesis}
\label{sec:Application}
We synthesize grasp poses of both hands based on the generated bimanual contact points by utilizing ContactGrasp~\cite{Brahmbhatt-2019ContactGraspFM}. ContactGrasp is a grasp optimizer for functional grasp generation from an object model and a contact map. It seeks a hand configuration coincided with a contact map by encouraging the hand to approach the contact points and discouraging to touch the other points.

In our implementation, the MANO~\cite{MANO-2017} hand model is adopted for optimization. First, we import MANO into GraspIt!~\cite{Miller-GraspitAV2004} to initialize the hand configuration. Then, we construct left and right-hand contact maps based on the contact points computed by BCPN, respectively. Finally, we compute the grasp configuration for each hand. Based on the right-hand contact map and the right MANO hand model, we optimize the right hand's grasp by ContactGrasp and choose the optimal grasp. Similarly, we can also obtain the optimal grasp of the left hand. After optimizing the grasps separately, we combine the grasps of bimanual hands as the final output of our application.
\section{Experimental Results}\label{sec:experimental}
In this section, we describe experiments to demonstrate the effectiveness, robustness and generalization of our proposed framework. We first introduce the implementation details of our network architecture and then perform ablation studies to verify the advantages of our network design. We also conduct qualitative and comparative experiments to evaluate the performance of the proposed network in terms of saliency map prediction, robustness to manually labeling and generalization to novel objects, and bimanual grasp synthesis. Finally, we assess the grasping capacity of our prediction on virtual synthetic object models in simulation and give the statistics for grasp success rates.
\begin{figure}[!t]
  \centering
  \begin{overpic}[width=1.0\columnwidth,tics=10]{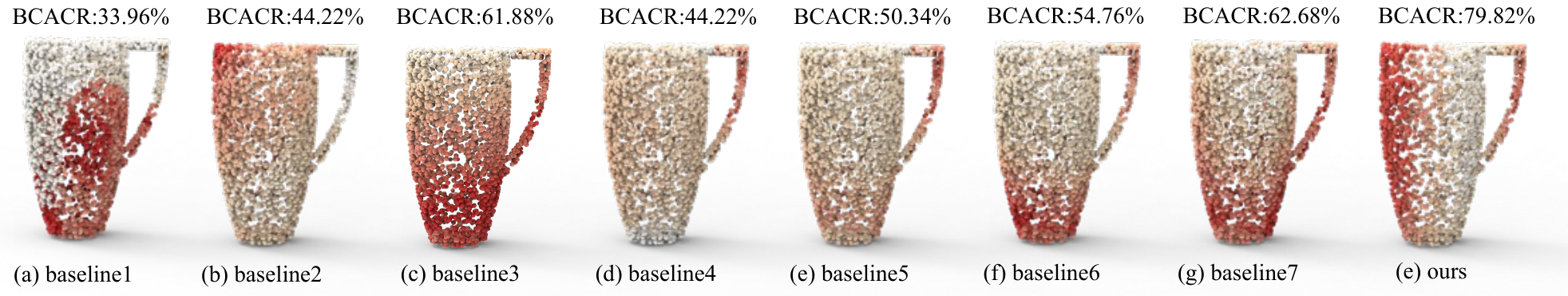}
  \end{overpic}
  \caption{Ablation Study. We compare our method to some possible alternatives to demonstrate the rationality of our model and loss function. We show the predicted saliency maps with various baseline variants and evaluate their plausibility by BCACR metric.
   }
   \label{fig:ablation_1}
   \vspace{-7pt}
\end{figure} 

\subsection{Implementation details}
\label{sec:Implementation}
We use the existing single-handed grasp saliency dataset~\cite{Lau-2016TactileMS} to train SSPN. All the $15$ household objects from this dataset, including Mug, Pot, Pan and Tackle, were held out as a training set. We also select 40 objects from ShapeNet object set~\cite{Chang2015ShapeNetAI} for these object categories as a testing set. The training data of saliency corresponding vectors are collected from human annotation and the input point clouds are captured from the object models present in~\cite{Lau-2016TactileMS} using the Blue noise Sampling method~\cite{xu2012blue}. We randomly sample 5000 surface points for each object during the training of BSPN and BCPN. In the process of iterative learning, we set up the hyperparameters $K=2000$ and $M=200$ for Mug and Tackle, and $K=1500$ and $M=250$ for Pot and Pan. The neural network is trained on a 2080Ti GPU.

\textbf{The single-handed grasp saliency dataset.} The dataset in~\cite{Lau-2016TactileMS} contains $15$ household object models and their single-handed grasp saliency maps. This dataset is collected by asking online users to choose a point which they prefer to grasp from a pair of points sampled on the model. The method~\cite{Lau-2016TactileMS} takes a learning-to-rank method to convert the ranking data to the single hand grasp saliency value for each point. 

\textbf{Contact points annotation.} We conducted a manual annotation of bimanual contact points on the object models present in~\cite{Lau-2016TactileMS}. Specifically, we invited $30$ users and ask them to utilize the CloudCompare software~\cite{Girardeau-2020} to mark the points where their both left and right hands will contact when grasping the model. Specifically, in the CloudCompare software, the user marks the contact area of the left and right hands on the surface of the model mesh with a rectangular frame. After that, we sample 5000 points on the mesh, and the points located in the contact area are the contact points.

\textbf{Saliency corresponding vector.} The ground truth of saliency corresponding vectors are built using the labels of contact points $L$. In practice, for a point $x_i$ labeled \textit{right-hand}, we compute the vector $y_j - x_i$ with each point $y_j$ labeled \textit{left-hand} as saliency corresponding vector. Thus, we can obtain a list of candidate vectors for $x_i$, and then we randomly select $1000$ vectors from them to build the ground truth set of saliency corresponding vectors. Similarly, we can also obtain the ground truth of saliency corresponding vectors for each point $y_i$ labeled \textit{left-hand}.

\textbf{Grasp configuration.} For grasp synthesis, we adopt MANO~\cite{MANO-2017} hand models as robotic hands to generate grasp configuration based on our bimanual saliency map. MANO hand is a parametric model of a human hand which grasp is 16-DoF. We import it into GraspIt!~\cite{Miller-GraspitAV2004} and make it available for ContactGrasp method~\cite{Brahmbhatt-2019ContactGraspFM}.

\subsection{Ablation study}
To justify the rationality of the design choice of each component in our network architecture, we compare our full pipeline against the following baseline variants.

\textbf{\emph{Baseline1:}} The network only includes $E_{\theta_1}-D_{\theta_3}$ component, and it is trained with the saliency corresponding vector predicted loss $L_c$. In the test phase, the saliency value of each point in the point cloud is updated to the saliency value of its corresponding point directed by the predicted saliency corresponding vector.

\textbf{\emph{Baseline2:}} The set of saliency corresponding vectors of all points labeled \textit{right-hand} is only adopted as the ground truth, instead of the original ground truth used in the training phase of our network.

\textbf{\emph{Baseline3:}} The network contains both $E_{\theta_1}-D_{\theta_3}$ and $E_{\theta_2}-D_{\theta_4}$ components, and it is trained without the proposed loss function $L_{a_2}$.

\textbf{\emph{Baseline4:}} The calculation of the loss function $L_{a_2}$ dose not contain $\psi(x)$ in the training phase.

\textbf{\emph{Baseline5:}} The network trained without the iterative learning strategy.

\textbf{\emph{Baseline6:}} The network is trained iteratively with correspondence loss adjustment loss $L_{c}$ but without physics-balance loss $L_{p}$.

\textbf{\emph{Baseline7:}} The network is trained with all loss functions and the physics-aware refinement module is removed during the testing phase.

To evaluate the plausibility of the predicted saliency maps in the ablation studies, we propose the bimanual contact annotation coverage rate (BCACR), measures the degree to which the bimanual grasp points derived from the saliency map are included within the bimanual grasp regions annotated by human observers. Specifically, we first count all annotated grasp contact points $C$ and then filter out grasp saliency points from the bimanual saliency map by a given threshold $\tau_c$, where $\tau_c=0.7$. We count the number of grasp saliency points $C_{cover}$ that have a bimanual-contact label of $1$ or $2$. Thus, BCACR is defined as the percentage of grasp saliency points which are covered by the bimanual contact annotated area among all annotated grasp contact points:
\begin{equation}
    \text{BCACR}=\frac{C_{cover}}{C} \times 100\%
\end{equation}
The higher value of BCACR, the more plausible the result of saliency map. Figure~\ref{fig:ablation_1} shows the results of ablation studies quantitatively and qualitatively, respectively.

As shown in Figure~\ref{fig:ablation_1}, our full method achieves better performance than the other baselines, which is fit to both human grasp preference and physical stability. Figure~\ref{fig:ablation_1}(a) indicates that the network cannot predict reasonable bimanual saliency results only from the original single grasp saliency and without the training process of saliency adjusted score. Furthermore, from Figure~\ref{fig:ablation_1}(b) we can see that the insufficient training data of the saliency corresponding vector has a negative impact on the prediction accuracy of saliency corresponding vector, leading to the result is undesirable. Figure~\ref{fig:ablation_1}(c-d) show that when the loss function $L_{a_2}$ is turned off in the network training stage, the $L_{a_1}$ becomes invalid and the saliency values of points that outside the labeled contact regions become increase, resulting in a large deviation with human grasp position selection. Notice that our iterative training strategy plays a critical role in learning bimanual saliency adjustment from original single hand saliency values. As illustrated in Figure~\ref{fig:ablation_1}(e), the initial saliency value is not updated in the training phase when the iteration strategy is not removed, which means that the grasp saliency of both hands is not adjusted simultaneously. Furthermore, Figure~\ref{fig:ablation_1}(f) shows that the results predicted by training without physics-balance loss only conform to manual annotations, which cannot satisfy physical balance. Figure~\ref{fig:ablation_1}(g) shows that for novel objects whose shapes is different significantly from objects in the training set, physically plausible results can’t be computed without physics-aware refinement, even if with the guidance of the physics-balance loss at the training stage.
\begin{figure}[ht]
  \centering
  \begin{overpic}[width=1.0\columnwidth,tics=10]{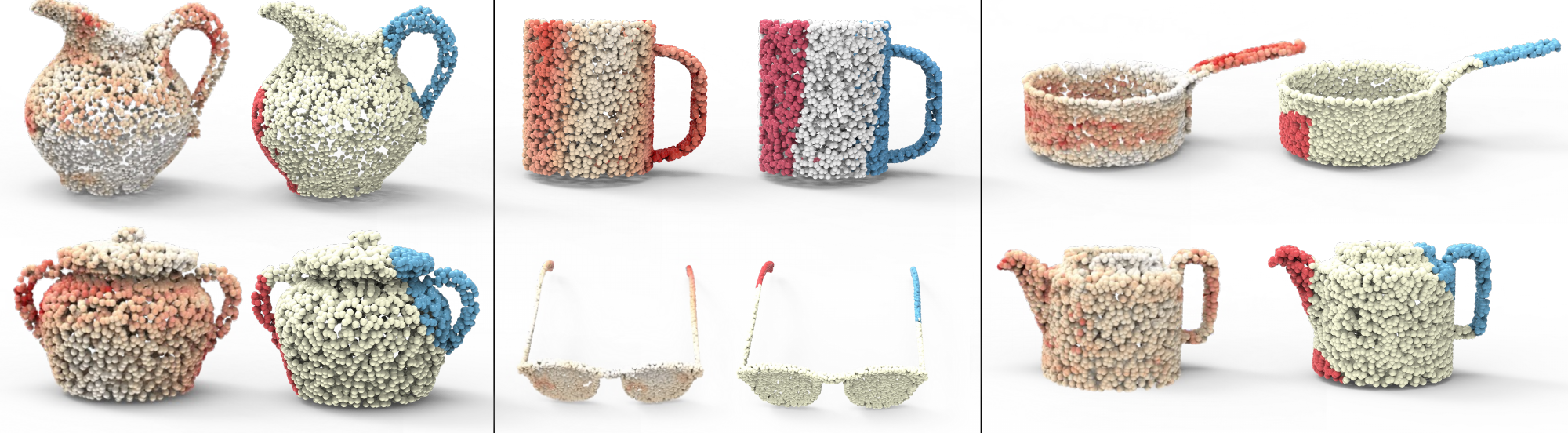}
  \end{overpic}
  \caption{Gallery of examples of grasp saliency maps and contact points predicted by our approach. The points colored by blue and red denote the predicted contact points of both hands, respectively.
   }
   \label{fig:qualitativeResults}
   \vspace{-7pt}
\end{figure} 

\subsection{Qualitative results and comparisons}
\textbf{Gallery.} Our network can perform better prediction performance of both grasp saliency map and contact positions of both hands. In Figure~\ref{fig:qualitativeResults}, we show a gallery of results of household objects predicted by our network. For each model, the contact points consist of a point set of left-hand contacts (denoted by red spheres) and a point set of right-hand contacts (denoted by blue spheres), which are predicted from the saliency map.

\begin{figure}[ht]
  \centering
  \begin{overpic}[width=1.0\columnwidth,tics=10]{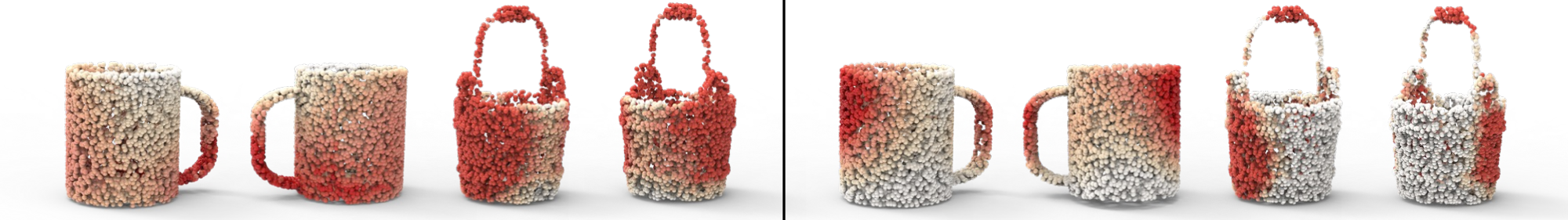}
  \put(6,-2){\small Left}
  \put(20,-2){\small Right}
  \put(31,-2){\small Left}
  \put(41,-2){\small Right}
  \put(56,-2){\small Left}
  \put(69.5,-2){\small Right}
  \put(80.5,-2){\small Left}
  \put(90.5,-2){\small Right}
  \end{overpic}
    \vspace{0.1cm}
  \caption{Our method is robust for interference in manual annotation. We show two object models with both left and right views respectively. The left columns are the results of training network neither using physics-balance loss nor physics-aware refinement module. The right columns are the results of training network both using physics-balance loss and physics-aware refinement module.
   }
   \label{fig:Robustness_to_manual_annotation}
   \vspace{-7pt}
\end{figure}
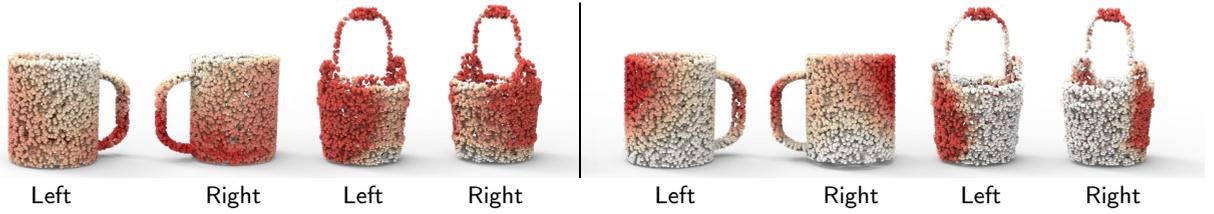 
Manual annotation of bimanual contact points is a key step in influencing the accuracy of the predicted corresponding vectors when training BSPN. In general, an accurate annotation helps to improve the capacity for bimanual saliency map prediction, but it is vulnerable to interference, such as human misoperation in annotation. To prove the robustness of our method to manual annotation, we show comparison results under annotation with noisy in Figure~\ref{fig:Robustness_to_manual_annotation}. Specifically, we randomly scramble the user’s bimanual contact point annotations on the object surface as disturbed annotation. The ratio of disturbed labeling to accurate labeling is 3:1. We train the models without and with physics-balance loss on the noisy dataset, respectively. Moreover, for the model which is trained without the physics-balance loss, we also do not use the physics-aware refinement module during testing. 

The results show that when the annotations are inaccurate, the model cannot predict a reasonable bimanual saliency map without using physics-balance loss and physics-aware refinement. However, when the physics-balance loss and the physics-aware refinement are added, our method not only relies on annotations but also considers physical criterion. Therefore, as illustrate in Figure~\ref{fig:Robustness_to_manual_annotation}, our method can predict physically plausible bimanual saliency map even if the manual annotations are under interference.
\begin{figure}[!t]
  \centering
  \begin{overpic}[width=1.0\columnwidth,tics=10]{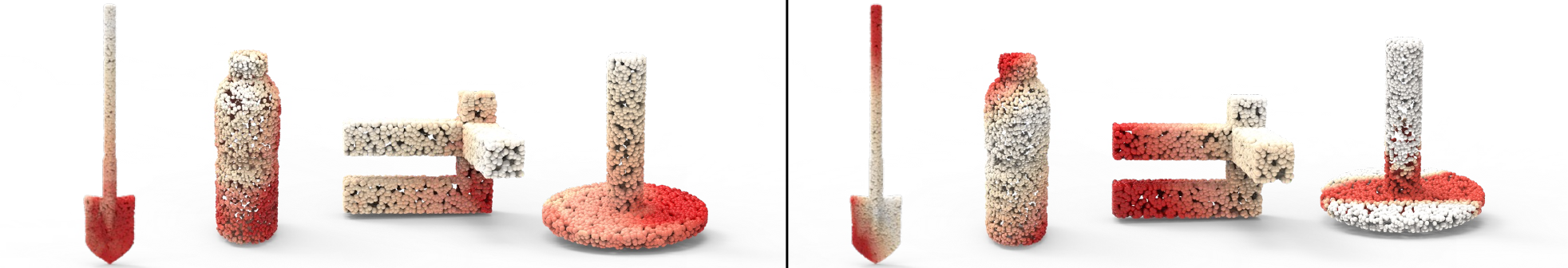}
  \end{overpic}
  \caption{Our approach achieves physically plausible results even if when human experience is not valid for novel objects. The two columns on the left show the saliency maps before physics-aware refinement. The two columns on the right show the results after physics-aware refinement.
   }
   \label{fig:unknown_object}
   \vspace{-7pt}
\end{figure} 

\textbf{Generalization to novel objects.}
Additionally, we also compared the performance of our approach on the physics-aware refinement to evaluate generalization to novel objects. To this end, we trained our model on the data of mug category and tested it on other object categories. Figure~\ref{fig:unknown_object} shows the qualitative results before and after executing the physics-aware refinement. We can see that the physics-aware refinement can notably improve generalization ability to novel objects, which is mainly due to it guarantees the physical balance for bimanual saliency map. In other word, our network is able to generate physically plausible result, even if when human experience is not valid for novel objects.
\begin{figure}[t!]
  \centering
  \begin{overpic}[width=1\columnwidth,tics=10]{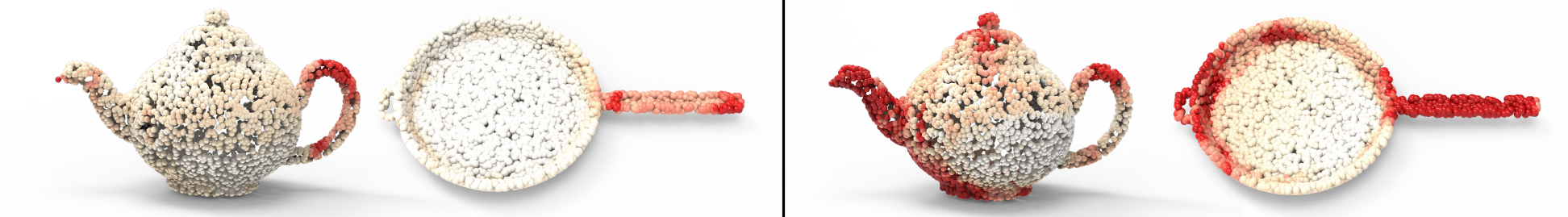}
  \end{overpic}

  \caption{Comparison of our method to the single-handed grasp saliency analysis method on pan and teapot models. The two columns on the left show saliency maps computed by the single-handed grasp saliency analysis method and the right columns show grasp saliency maps computed by our method.
   }
   \label{fig:qualitativeResults_3}
   \vspace{-7pt}
\end{figure} 

\textbf{Comparison with the single-handed grasp saliency method.} To verify the necessary of jointly learning the grasping saliency of both hands, we compare with a single-handed saliency analysis method SSPN~\cite{Zhang-20203DGS}. Compared with SSPN, we consider the bimanual grasp saliency prediction as a stacking problem of grasp saliency of left and right hands. We use SSPN to compute both the right-handed and left-handed grasp saliency map. Specifically, we build a small dataset to train SSPN network so as to compute the left-handed grasp saliency map. We ask users to label contact points in the point cloud where their left hands prefer to grasp when grasping with single hand. Thus, we collect a total of 30 users' annotations and compute the probability that each point in the point cloud is selected by the user as a left-handed saliency value. Using this small dataset, we train SSPN to obtain the left-handed grasp saliency map. Therefore, we combine the right-handed grasp saliency map with the left-handed grasp saliency map and normalize it as the final result. Figure~\ref{fig:qualitativeResults_3} shows the comparison results both on pan and teapot models. Two columns on the left in Figure~\ref{fig:qualitativeResults_3} show the saliency maps computed by the SSPN and the right show the results computed by our method. In the Figure~\ref{fig:qualitativeResults_3}, the bimanual saliency map computed by SSPN is the same as that of the single hand. It proves that grasping objects with the left or right hand in isolation that are far from the human preference of grasping objects cooperatively with both hands. In contrast, our method jointly predicts the saliency values of both hands and can obtain more reasonable saliency maps.
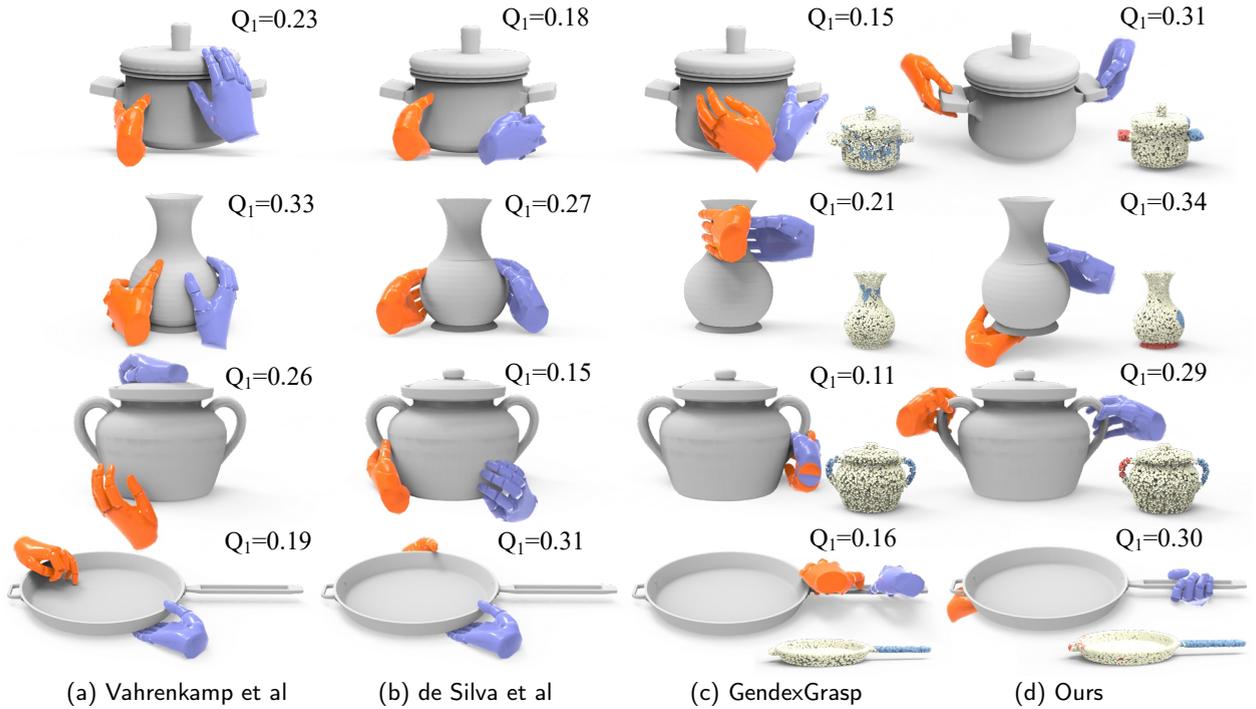
\begin{figure}[t!]
  \centering
  \begin{overpic}[width=1\columnwidth,tics=10]{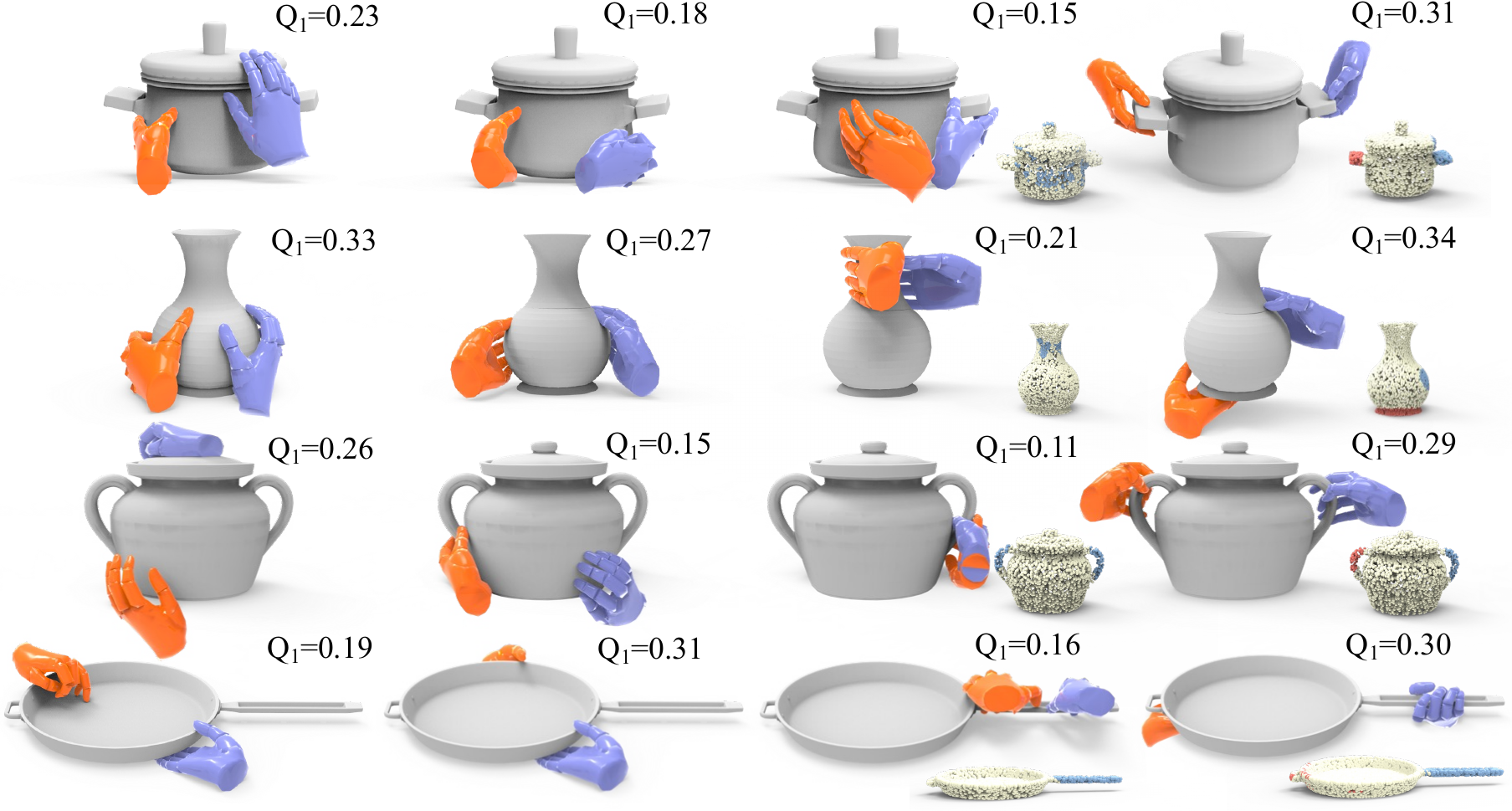}
  \put(5,-2){\small (a) Vahrenkamp et al}
  \put(30,-2){\small (b) de Silva et al}
  \put(55,-2){\small (c) GendexGrasp}
  \put(81,-2){\small (d) Ours}
  \end{overpic}
  \vspace{0.1cm}
  \caption{Bimanual grasps of the pot, vase, kitchenpot and pan models. The first two column (a) and (b) show the results generated by~\cite{Vahrenkamp-2011BimanualGP,RojasdeSilva-2016GraspingBO}. The column (c) show results computed by GendexGrasp~\cite{li2022gendexgrasp}. The last column (d) show the bimanual grasp computed by method~\cite{Brahmbhatt-2019ContactGraspFM} from our predicted contact points of both hands.
   }
   \label{fig:qualitativeResults_2}
   \vspace{-7pt}
\end{figure} 

\textbf{Comparison with bimanual grasp synthesis.} The existing literatures on bimanual grasp planning only consider force-closure contact points on the object surface. Therefore, it is difficult for them to synthesize grasp that is amenable to human grasp experience. We compare our method with sampling-based grasp optimization proposed by methods~\cite{Vahrenkamp-2011BimanualGP,RojasdeSilva-2016GraspingBO} and contact-based grasp synthesis proposed by GendexGrasp method~\cite{li2022gendexgrasp}, and evaluate the grasp quality by computing the generalized Q1 analytic measure for grasp stability proposed by method~\cite{2020Deep}. Figure~\ref{fig:qualitativeResults_2} shows the comparison of bimanual grasp synthesis for four target objects. The first two column (a) and (b) show the results generated by~\cite{Vahrenkamp-2011BimanualGP,RojasdeSilva-2016GraspingBO}. The third column (c) show results computed by GendexGrasp~\cite{li2022gendexgrasp}. The last column (d) in Figure~\ref{fig:qualitativeResults_2} show the bimanual grasp computed by method~\cite{Brahmbhatt-2019ContactGraspFM} from our predicted contact points of both hands. It demonstrates that our approach is helpful in generating grasps that are notably closer to the human grasp preference than the physics-based analytic method.
\begin{figure}[t!]
  \centering
  \begin{overpic}[width=1.0\columnwidth,tics=10]{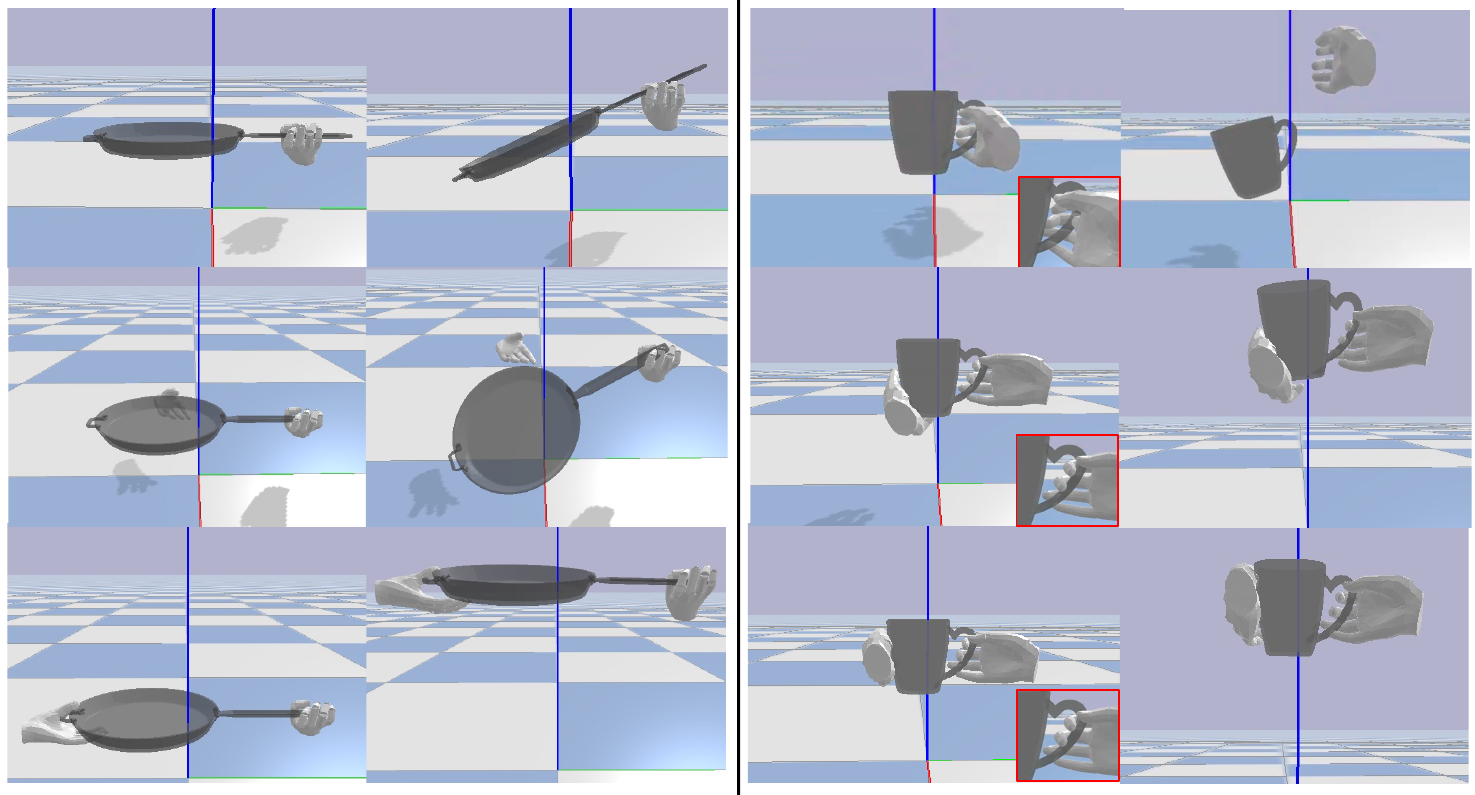}
  \end{overpic}

  \caption{Progressive demonstrations of grasping objects with single dexterous hand and both dexterous hands in simulation. The first row show the single-handed grasping results generated by method~\cite{Brahmbhatt-2019ContactGraspFM}. The last two row are the stages before and after using the physics-balance loss and physics-aware refinement module.
}
   \label{fig:simulation}
   \vspace{-7pt}
\end{figure} 
\subsection{Simulation experiments with dexterous hands}
We further carry out our experiments in real-time physical simulator~\cite{coumans-2019} and quantitatively evaluate our approach with MANO hands and the single-handed grasp planning method~\cite{Brahmbhatt-2019ContactGraspFM} in terms of grasping success rate. 80 shape models from 8 object categories (including 4 novel object categories) are selected for evaluation. Firstly, we compute bimanual grasp candidates for each object and select the optimal one according to the metric of ContactGrasp. As illustrated in Figure~\ref{fig:simulation}, we test if a grasp is successful in the simulation environment by applying the generated bimanual grasp and measuring the movement of object. We test each grasp by moving the hands upwards to 15 cm above the ground and evaluate if the object tilts or drops out during translation. We repeat this process for each category of object and measure their grasping success rates. 

We can see that if the physics-balance loss and physics-aware refinement module are not used, the grasp will tilt or even fall, but both relying on both the human experience and the physical criterion, the result achieves more stable and reasonable. We also observe that when the object size is larger relative to one hand, such as the pan in Figure~\ref{fig:simulation}, it's very difficult to perform successful grasp either by grasping the handle or body of the pan with the single hand. However, the bimanual grasping is more stable than single-handed grasping.

As table~\ref{tab:GraspSuccessfulRate} illustrated, the generated grasps from our saliency map, on average, achieves 92.5\% success rate for all object categories, compared to an average success rate of only 70.62\% for single-handed grasping generated by the method~\cite{Brahmbhatt-2019ContactGraspFM}, which suggests that the proposed method can not only improve grasp quality in bimanual grasping, but also has notable superiority than single-handed grasping.

\begin{table}[ht]
\centering
\caption{Grasping success rate.}
\resizebox{1\columnwidth}{!}
{
\begin{tabular}{|c|c|c|c|c|c|c|c|c|}
\hline 
Target Object's Category & Mug & Pot & Pan & Tool & Kitchen Pot & Vase & Kettle & CAD Models \\ \hline
Single-handed Grasping Success Rate & 80\% & 70\% & 75\% & 70\% & 50\% & 80\% & 70\% & 70\% \\ \hline
Ours Grasping Success Rate &100\% &100\% &100\% &80\% &90\% &100\% &90\% &80\%      \\ \hline
\end{tabular}
}
\label{tab:GraspSuccessfulRate}
\end{table}

\section{Conclusion and Discussion}
In this work, we present a physics-aware iterative learning and prediction of saliency map for bimanual grasp planning that can generate bimanual contact points that are physically plausible for grasping balance and amenable to human grasping preferences. Without relying on large training data of bimanual grasp annotations, our method can predict bimanual grasp saliency from single-handed grasp saliency, through learning saliency corresponding vectors that conduct the relationships of grasp contacts between both hands by a small amount of bimanual grasp annotations. This facilitates bimanual grasp synthesis for household objects. Extensive experimental results on dexterous hands exhibit the effectiveness of our method.

Our current solution still has some limitations. We leverage the optimizer to synthesize grasps of both hands guided by bimanual saliency map. However, the grasp optimizer cannot simultaneously generate grasps of both hands, which is due to it needs a more complex grasp initialization. As a result, we have to synthesize the grasp of each hand separately. Hence, in the future work, we plan to integrate the bimanual grasp generation into the network for performing bimanual grasp planning in end-to-end fashion.
\section*{Acknowledgement}
We thank the anonymous reviewers for their valuable comments. This work was supported by NSFC programs (62272082), Emerging Interdisciplinary Cultivation Project of Jiangxi Academy of Sciences (2022YXXJC0101).

\bibliographystyle{cas-model2-names}
\bibliography{reference}

\end{document}